\documentclass[runningheads]{llncs}

\usepackage{eccv}
\usepackage{pifont}
\usepackage{eccvabbrv}
\usepackage{graphicx}
\usepackage{booktabs}
\usepackage[accsupp]{axessibility}

\newcommand{\our}{$S^{2}$-FracMix}
\usepackage{graphicx}  
\usepackage{wrapfig}    
\usepackage{xcolor}     

\usepackage[ruled,vlined]{algorithm2e}
\definecolor{darkgreen}{rgb}{0.0, 0.5, 0.0}
\newcommand{\gain}[1]{\textbf{\textcolor{darkgreen}{#1}}}

\usepackage{booktabs} 

\usepackage{multirow}
\usepackage{xcolor}
\usepackage[table]{xcolor} 
\usepackage[table, dvipsnames]{xcolor} 

\usepackage{hyperref}

\usepackage{orcidlink}

\begin{document}

\title{$S^{2}$-FracMix: Label-Preserving Self-Saliency Mixup Augmentation}

\titlerunning{$S^{2}$-FracMix}

\author{
Khawar Islam\inst{1}\orcidlink{0000-0003-0674-8637}\thanks{Corresponding author: \email{islam.k@unimelb.edu.au}} \and
Arif Mahmood\inst{2}\orcidlink{0000-0001-5986-9876} \and
Xin Jin\inst{3}\orcidlink{0009-0005-0983-6853} \and
Naveed Akhtar\inst{1}\orcidlink{0000-0003-3406-673X}
}

\authorrunning{K.~Islam et al.}

\institute{
The University of Melbourne, Australia
\and
Information Technology University, Pakistan
\and
Westlake University, China
}

\maketitle

\begin{abstract}
Data augmentation is known to improve  generalization of deep visual models. Recent  methods favor mixup strategies that generate interpolated samples to improve model performance. However, these techniques not only incur significant computational overhead, they also lead to semantic disruption of augmentation data due to cross-sample mixing. We first propose Self-Saliency ($S^2$) Mixup, which constructs challenging yet label-consistent samples by extracting multi-scale salient patches and reinserting them into non-salient regions of the same image. This promotes scale-invariant feature learning while avoiding cross-sample interference. To further enhance model robustness, we introduce FracMix, a mixing scheme that injects self-similarity patterns into salient regions using adaptive ratios. Collectively, our unified framework, $S^{2}$-FracMix, enables simultaneous learning from fractal and non-fractal structures within a single image, yielding a targeted and structurally coherent augmentation strategy. We theoretically analyze the advantage of our technique, and empirically establish its superiority over the existing methods by achieving state-of-the-art performance in extensive evaluation with seven benchmarks across classification (coarse and fine-grained), robustness, calibration, object detection, and transfer learning tasks. Project page is available at \href{https://fracmix-data-augmentation.github.io/}{fracmix-data-augmentation.github.io}

\end{abstract}

\vspace{-15pt}
\section{Introduction}
\label{sec:intro}
\vspace{-5pt}
The rapid growth in the scale and representational capacity of Deep Neural Networks (DNNs) has enabled modern models to effectively memorize training data \cite{zhang2018mixup, cao2024survey, carratino2022mixup, won2025effects}. While this capacity contributes to strong empirical performance, it also exacerbates overfitting, thereby widening the generalization gap. To address this challenge, data augmentation has emerged as a fundamental research direction \cite{kang2023guidedmixup, kim2020puzzle, kim2021co, parast2026hsfm, parast2025ddb}. Data augmentation strategies have been widely adopted across diverse vision tasks, including image classification \cite{qin2025sumix, chen2022transmix}, object detection \cite{zoph2020learning}, and segmentation \cite{ghiasi2021simple, jin2025mergemix}. By enriching training distributions, these methods improve robustness to unseen data, mitigate model collapse \cite{kang2023guidedmixup, xiao2023token, wang2024enhance}, and enhance resilience to distribution shifts \cite{pinto2022using, jin2024survey, parast2025ddb}.
\par
\begin{figure*}[t]
    \centering
    \includegraphics[width=1.0\linewidth]{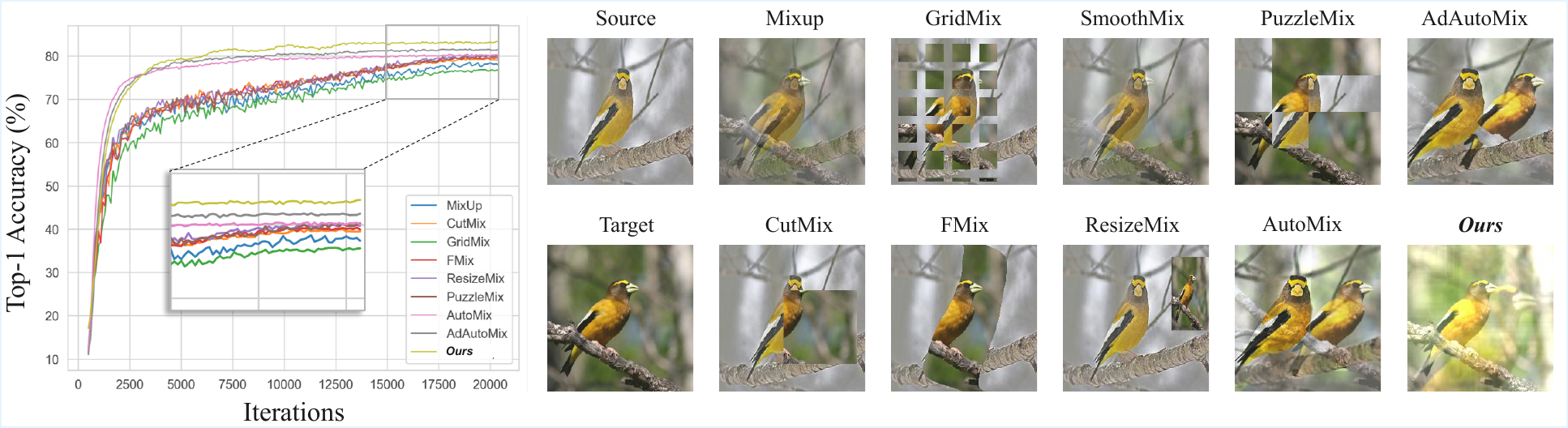}
    \vspace{-9pt}
    \caption{ \textcolor{black}{(Left) Performance of ResNet-18 trained  with various augmentation methods for 200 epochs on CIFAR100.} (Right) Representative augmentation samples created by different methods. The samples get constructed with source and target images.}
    \label{fig:2}
    \vspace{-9pt}
\end{figure*}
\begin{wrapfigure}{r}{0.45\textwidth}
    \vspace{-15pt}
    \includegraphics[width=0.45\textwidth]{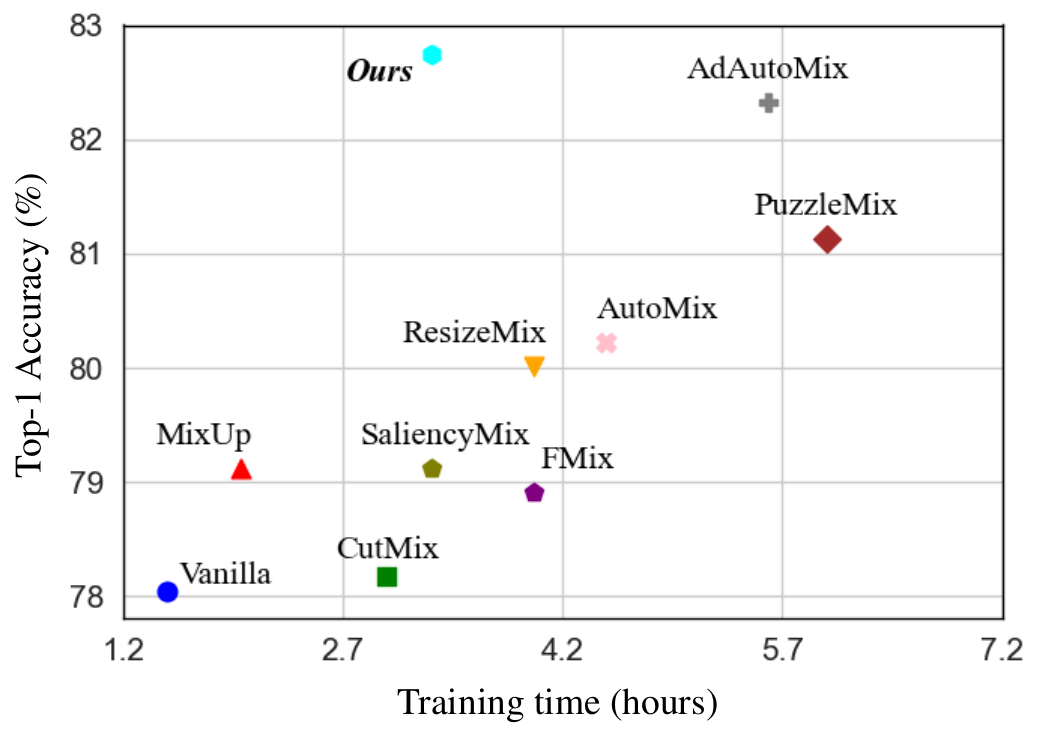}
    \caption{Comparisons of total training time \textit{vs.} top-1 accuracy of \texttt{ResNet-18} on CIFAR-100 dataset with RTX 3090.}
    \label{fig:1}
    \vspace{-10pt}
\end{wrapfigure}

A central objective of modern augmentation techniques is to increase sample diversity and robustness while preserving the structural and semantic integrity of the underlying data \cite{jin2024survey, huang2023ipmix, han2022cropmix, parast2025ddb, parast2025ghost, hendrycks2020augmix, jin2025mergemix}. Equally important is their practical applicability, which requires a careful balance between performance gains and computational overhead \cite{kim2021co, kim2020puzzle,islam2025context,islam2024genmix}. In this work, we introduce \our{}, a computationally efficient augmentation framework that promotes greater diversity and structural complexity in augmented samples, thereby improving generalization without incurring substantial overhead (see Fig.~\ref{fig:2} \& \ref{fig:1}).


Within the broader task-agnostic \textit{mixup} paradigm, samples are typically generated   using random pairs of training instances \cite{zhang2018mixup}. Representative methods such as CutMix~\cite{yun2019cutmix}, Manifold Mixup~\cite{verma2019manifold}, AlignMixup~\cite{venkataramanan2022alignmixup}, and ResizeMix~\cite{qin2020resizemix} adopt different mixing strategies to construct synthetic training examples. 
Since these techniques combine random pairs of images, they fall short on preserving semantically salient image regions, potentially leading to suboptimal structural consistency.
To address this limitation, saliency-guided methods - including SaliencyMix \cite{uddin2020saliencymix}, PuzzleMix \cite{kim2020puzzle}, Co-Mixup \cite{kim2021co}, and GuidedMixup \cite{kang2023guidedmixup} - were introduced to prioritize informative regions during mixing. While effective, these approaches typically incur substantial computational overhead, increasing training time and demanding high-end computational resources \cite{kim2020puzzle, kim2021co}. 
Other methods such as AutoMix \cite{liu2022automix} and AdAutoMix \cite{qinadversarial}  attempt to automatically learn optimal mixing strategies and label assignments. However, they remain less effective for Transformer-based architectures, besides their heavy compute overhead. 
Similarly, fractal-based techniques \cite{islam2024diffusemix, huang2023ipmix, hendrycks2022pixmix} end up  disrupting essential image content while inducing undesirable distribution shifts, which compromises model robustness.

Addressing these challenges, we propose \our{}, a unified augmentation framework comprising two complementary components: Self-Saliency ($S^2$) and \textit{FracMix}. The $S^2$ component extracts multi-scale, saliency-guided patches from an input image, applies diverse transformations to these patches, and reintegrates them into the same image through a controlled mixing operation. By operating within a single image, $S^2$ avoids cross-sample semantic interference while promoting structural diversity and scale-invariant representation learning.
Building upon this foundation, \textit{FracMix} introduces targeted fractal-based augmentation. In contrast to methods such as DiffuseMix \cite{islam2024diffusemix}, which diffuses fractal textures across the entire image, \textit{FracMix} confines self-similar fractal injection to the salient regions identified by $S^2$. This targeted design increases structural complexity while preserving the clean contextual background. Consequently, each augmented sample simultaneously contains fractal and non-fractal structures, enhancing diversity without inducing undesirable distribution shifts. 

Rather than relying on a fixed mixing heuristic, \our{} incorporates multiple mixing modes that are randomly selected for each training instance. This high-level stochastic mixing strategy further diversifies the training distribution, enabling the model to learn more robust object representations and generalize effectively to unseen data. Our extensive evaluations establish  that \our{} consistently outperforms state-of-the-art augmentation baselines across seven datasets and nine competitive methods. Our approach achieves superior performance in clean accuracy, adversarial robustness, recognition under occlusion, and a broad range of downstream scenarios. The main contributions of this work are summarized as follows:

\begin{itemize}

\item We introduce Self-Saliency ($S^{2}$) mixing, a multi-scale, saliency-guided augmentation strategy that extracts informative patches, applies diverse transformations, and reintegrates them into the same image. 

\item We propose \textit{FracMix}, a targeted fractal-based mixing mechanism that injects self-similar structures exclusively within salient regions, increasing structural complexity while preserving contextual integrity and data fidelity.

\item We develop a unified framework with $S^{2}$ and \textit{FracMix}, termed $S^{2}$-FracMix, and extend it with a proposed concept of multi-mode mixing. We also theoretically analyze the key strength of our framework.

\item Through comprehensive evaluations on seven datasets and comparisons with nine state-of-the-art methods, we demonstrate consistent improvements across general and fine-grained classification, object detection, transfer learning, self-supervised learning, calibration, and few-shot learning.

\end{itemize}

\vspace{-3pt}
\section{Related Work}
\label{relatedWork}
\vspace{-2.5pt}

\subsection{Mixup Augmentation}
Mixup-based augmentation generates  diverse training samples through interpolation strategies \cite{lee2020smoothmix, yang2022recursivemix, hong2021stylemix}. Manifold Mixup \cite{verma2019manifold} extends linear interpolation to hidden representations, encouraging smoother decision boundaries in latent space. CutMix \cite{yun2019cutmix} replaces rectangular regions between images to promote occlusion-aware learning. Subsequent variants such as FMix \cite{harris2020fmix} and GridMix \cite{baek2021gridmix} adopt structured masking strategies, while ResizeMix \cite{qin2020resizemix} resizes and overlays patches to introduce scale-aware transformations. SnapMix \cite{huang2021snapmix} improves fine-grained recognition by proportionally mixing semantically relevant regions. More recently, Decoupled Mixup \cite{liu2024harnessing} proposed an efficient objective with a decoupled regularizer that leverages hard mixed samples to mine discriminative features. Collectively, these methods enhance generalization by synthesizing new samples, yet they largely rely on inter-image mixing without explicitly preserving semantic saliency.

\subsection{Automated Mixup Augmentation}
Automated mixup approaches balance mixing strategy design with optimization complexity, addressing the disconnect between the mixing heuristics and downstream training objectives. AutoMix \cite{liu2022automix} jointly optimizes sample generation and classification to  produce task-relevant mixed examples. Related directions include adversarial data augmentation \cite{zhao2020maximum} and GAN-based methods \cite{antoniou2017data}, which seek to automate augmentation through learned generators. Adversarial MixUp \cite{qinadversarial, won2025understanding} further tackles domain shift by synthesizing mixed samples for adaptation. While effective, these approaches often introduce additional training complexity and computational overhead. Their generalizability across network architectures also remains restricted. 

\subsection{Saliency-driven Mixup Augmentation.}
Saliency-guided methods prioritize discriminative regions during mixing to preserve semantic integrity. SaliencyMix \cite{uddin2020saliencymix} and Attentive-CutMix \cite{walawalkar2020attentive} combine the most informative regions from source and target images. PuzzleMix \cite{kim2020puzzle} redistributes patches based on saliency and local statistics, and Co-Mixup \cite{kim2021co} extends this formulation to multi-image mixing with supermodular diversity constraints. SAMix \cite{li2021boosting} decomposes mixup objectives into locally emphasized and globally constrained terms to enable adaptive mixing. SalfMix \cite{choi2021salfmix} generates self-mixed samples by transferring salient regions into less salient areas within the same image. GuidedMixup \cite{kang2023guidedmixup} focuses on critical local features using spectral residual-based saliency.
Despite these advances, most saliency-driven methods rely on transferring salient regions across different images, which  introduces semantic inconsistencies. 

\vspace{-3pt}
\section{Proposed Method}
\label{sec:PM}
\vspace{-3pt}
\subsection{Overview}

The motivation behind the proposed \our{} is to directly encode self-contained multi-scale saliency-guided augmentation. We preserve object saliency while promoting structural diversity. Unlike prior works, we take a direct approach for the detection of salient regions via spectral residual method \cite{hou2007saliency}, which is then used to guide patch extractions at various scales.

\begin{wrapfigure}{r}{0.50\textwidth}
    \vspace{-40pt}
    \includegraphics[width=0.48\textwidth]{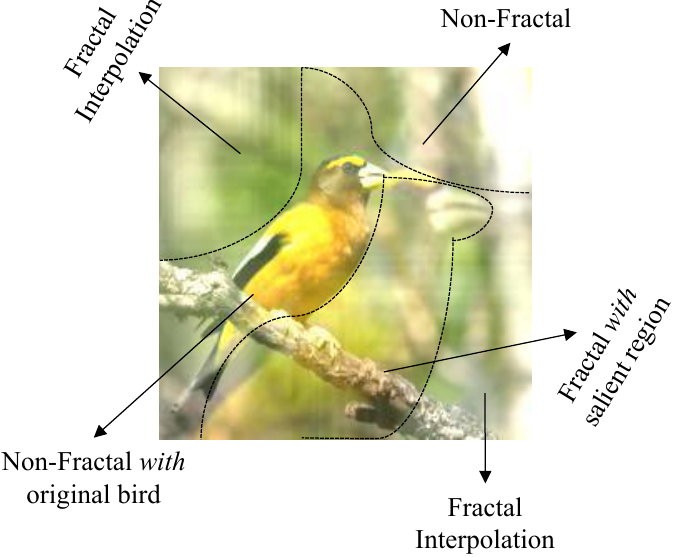}
    \caption{Augmentation samples of  \our{} are intricate composition of fractal and non-fractal patterns interpolated at multiple scales, all while retaining original image semantics.}
    \label{fig:BMS}
    \vspace{-3pt}
\end{wrapfigure}

These patches are transformed using rotation and blurring and mixed with the original image at non-salient random positions. Thus, \our{} explicitly encodes scale-invariant representation learning while preserving semantic integrity. As shown in Algorithm \ref{alg:multiscale_mixup} and illustrated in Fig. \ref{fig:BMS}, these multiscale patches are mixed with the original image, ensuring that important information highlighted through saliency is retained, while background or less discriminative areas are altered, to strengthen robust feature learning. 

\subsection{Self Saliency ($S^2$) Mixing}
\label{sec:self-saliency_method}
Let $\mathcal{D} = \{(I_i, y_i)\}_{i=1}^N$ represent the training dataset, where $I_i \in \mathbb{R}^{c \times h \times w}$ is an input image with $c$ channels, height $h$, and width $w$, and $y_i$ is its corresponding one-hot encoded label. Self Saliency mixup generates an augmented image $\tilde{I}_i$ (with $\tilde{y}_i = y_i$) as follows. Saliency maps $S_i \in \mathbb{R}^{1 \times h \times w}$ are computed to highlight regions critical to the predictions of model as:
\begin{equation}
    S_i = f(I_i),
\end{equation}
where $f(\cdot)$ is the saliency detection method. The saliency maps guide the selection and transformation of patches. Patches are extracted from the input image $I_i$ at $2$ patch scales $\mathcal{P} = \{(h/2,w/2),(h/4,w/4)\}$. For a patch $P_k \in \mathcal{P}$, the top left position $(x_k, y_k)$ is sampled uniformly
$ x_k \sim \text{Uniform}(0, w - w_k)$, and $y_k \sim \text{Uniform}(0, h - h_k)$.
The patch is then extracted as:
$ P_k = I_i[x_k:x_k + w_k, y_k:y_k + h_k]$.
Let $S_k$ represent the saliency mask for the corresponding patch, defined as:
$ S_k = S_i[ x_k:x_k + w_k, y_k:y_k + h_k]$.
$S_k$ is normalized to $[0,1]$ and a random threshold $t \sim \text{Uniform}(0.5, 1.0)$ is applied to obtain the binarized mask $\tilde{S}_k$. The patch is accepted only if the salient area ratio meets $\sum \tilde{S}_k / (h_k w_k) \ge (1-t)$; otherwise it is rejected. Each accepted patch is first blended with a randomly selected fractal image $F$ as described in Section~3.3 to obtain $P_k^f$. The transformation $T_k(P_k^f, \tilde{S}_k)$ is applied as
\begin{equation}
    T_k = R(P_k^f, \theta) \cdot \tilde{S}_k + B(P_k^f) \cdot (1 - \tilde{S}_k),
    \label{equ:three}
\end{equation}
where $R(P_k^f, \theta)$ applies a random rotation $\theta \sim \text{Uniform}(-30^\circ, 30^\circ)$ and $B(\cdot)$ applies Gaussian blurring. The transformed patch is resized back to the original image dimensions.
\begin{equation}
    R_k = \text{Resize}(T_k, (h, w))
    \label{euq:eq4}
\end{equation}
These resized patches are mixed into the original image using a weighted sum
\begin{equation}
    \tilde{I}_i = \alpha I_i + (1-\alpha) \sum_{k=1}^{n_p} \lambda_k R_k,
    \label{euq:equ8}
\end{equation}
where $\lambda_k=1/n_p$ are mixing weights with $\sum_{k=1}^{n_p} \lambda_k=1$, $n_p$ is the number of accepted patches, and $0 \le \alpha \le 1$ is a uniform random variable. Thus, $S^2$ drives learning models to handle a range of spatial transformations without complex mask optimization procedures as used in previous methods such as SaliencyMix \cite{uddin2020saliencymix}, PuzzleMix \cite{kim2020puzzle}, and Co-Mixup \cite{kim2021co}. As a result, our method remains computationally efficient yet highly diverse, synthesizing effective mixing modes that preserve semantic cues.

\begin{algorithm}[t]
\caption{\our{} Algorithm}
\SetAlgoSkip{smallskip}
\SetInd{0.25em}{0.5em}
\setlength{\algomargin}{8pt}
\label{alg:multiscale_mixup}
\small
\SetAlgoLined
\DontPrintSemicolon

\textbf{Require:} $\mathcal{I}_b = \{I_i, \textbf{y}_i\}_{i=1}^b$, $\mathcal{P}$, $\mathcal{F}$, $\lambda$ \\
\textbf{Ensure:} $\widetilde{\mathcal{I}}_b = \{\tilde{I}_i, {\textbf{y}}_i\}_{i=1}^{b}$: Augmented batch with labels \\

\ForEach{image $(I_i, y_i) \in \mathcal{I}_b$}{
    $S_i \leftarrow \texttt{SaliencyMap}(I_i)$, $P_m \leftarrow \texttt{zeros}(w,h)$, $n_k \leftarrow 0$\;

    \ForEach{patch scale $\mathbf{p}_k(h_k, w_k) \in \mathcal{P}$}{
        $x_k \leftarrow \texttt{uniform}(1, h-h_k)$,
        $y_k \leftarrow \texttt{uniform}(1, w-w_k)$\;

        Image patch: $P_k \leftarrow I_i[ x_k : x_k+h_k,\, y_k : y_k+w_k]$\;
        Saliency patch: $S_k \leftarrow S_i[ x_k : x_k+h_k,\, y_k : y_k+w_k]$\;

        $S_k \leftarrow (S_k-\texttt{min}(S_k))/(\texttt{max}(S_k)-\texttt{min}(S_k))$\;

        $t \leftarrow \texttt{uniform}(0.50,1.0)$, $\tilde{S}_k \leftarrow \texttt{threshold}(S_k,t)$\;

        \If{\texttt{Sum} $(\tilde{S}_k)/(w_k h_k) < (1-t)$}{
            \texttt{reject} $P_k$\;
        }
        \Else{
            $F \leftarrow \texttt{uniform}(\mathcal{F})$, $\theta \leftarrow \texttt{uniform}(-30,30)$\;
            Fractal blending: $P_k^f = \lambda F + (1-\lambda) P_k$\;
            Transform: $T_k \leftarrow \text{R}(P_k^f,\theta)\cdot \tilde{S}_k + \text{B}(P_k^f)\cdot (1-\tilde{S}_k)$\;
            Resize patch: $R_k \leftarrow \texttt{Resize}(T_k,(h,w))$\;
            Mixed patch: $P_m \leftarrow P_m + R_k$\;
            $n_k \leftarrow n_k + 1$\;
        }
    }

    $\alpha \leftarrow \texttt{uniform}(0,1)$\;
    Augment image: $\tilde{I}_i \leftarrow \alpha I_i + \frac{1-\alpha}{n_k} P_m$\;
    $\widetilde{\mathcal{I}}_b \leftarrow \widetilde{\mathcal{I}}_b \cup \{(\tilde{I}_i,\mathbf{y}_i)\}$\;
}

\textbf{return} $\widetilde{\mathcal{I}}_b$
\end{algorithm}

\subsection{FracMix \& High-level Mixing Modes}
\label{sub:fractal_minmodes}
In traditional methods, fractals are blended with the whole input image. In contrast, we propose fractal blending only inside the salient patches selected by the Self-Saliency (S²) procedure. We maintain a fixed library of $\mathcal{F}$ of fractal patterns constructed before training  for each accepted patch $P_k$, we sample $F \leftarrow \texttt{uniform}(\mathcal{F})$ and blended it with $P_k$ using factor $\lambda$ as
\begin{equation}
    P_k^f = \lambda F + (1-\lambda) P_k,
    \label{equ:six}
\end{equation}
In practice, we select $\lambda$ empirically via a small sweep on a held-out validation split (see Sec. \ref{main_paper:Hyperparameters}, Fig. \ref{fig:hyparameter}), and use $\lambda$ =0.20 as the default setting in all experiments. The resulting fractal-enriched patch $P_k^f$ then undergoes the selective transformation $T_k$ (Eq.~\ref{equ:three}) followed by resizing to full resolution (Eq.~\ref{euq:eq4}) to obtain $R_k$, and mixing in Eq.~(\ref{euq:equ8}).


It is notable that existing approaches generally employ a {single} mixing strategy throughout training, such as Mixup \cite{guo2019mixup}, CutMix \cite{yun2019cutmix}, ResizeMix \cite{qin2020resizemix}, PuzzleMix \cite{kim2020puzzle}, and GuidedMixup \cite{kang2023guidedmixup}. 
We empirically observed that restricting the model to only one mode of low-level mixing limits the diversity of supervisory signals,  compromising model performance. Therefore, we propose to incorporate \emph{multiple low-level modes of mixing} within the training pipeline. Specifically, we allow mixing  computationally efficient mechanisms of Mixup   \cite{guo2019mixup}, CutMix \cite{yun2019cutmix}, and ResizeMix \cite{qin2020resizemix} jointly at high-level with our \our{} technique to further the performance gain. The process behind this mode mixing randomly selects the mechanisms to encourage complementary regularization effects and expose the model to a richer variety of mixed inputs, ultimately improving model generalization. We thoroughly analyze this in Sec. \ref{sec:Ablation}.

\vspace{-2pt}

\section{Theoretical Analysis}
\label{lab:proof_analysis}

Our self-saliency mixing forces high-saliency feature reconstruction in low-saliency context. This can be viewed as effectively adding a structured auxiliary loss that encourages the minimal sufficient statistic to be distributed across the entire image rather than concentrated in a few dominant regions, which improves robustness to local perturbations. Below, we provide a theoretical analysis of this property of our method.

Let $(x,y)\sim\mathcal{D}$ denote a training sample drawn from the data distribution, where $x\in[0,1]^{C\times H\times W}$ is an input image with $C$ channels and spatial resolution $H\times W$, and $y$ is its ground-truth label. Let $h_\omega$ denote the predictor, and let $\ell(\cdot,\cdot)$ denote the training loss. For notational convenience, we define
\begin{equation}
g(z):=\ell(h_\omega(z),y),
\end{equation}
i.e., $g$ is the loss viewed as a function of the input $z$ while keeping the label $y$ fixed. We denote by $\theta\sim\Pi$ the random augmentation parameters, where $\theta$ collects all randomness in the augmentation process (sampled patch scale and location, saliency mask, rotation angle, fractal sample, and the mixing coefficient $\alpha$). We define the augmentation operator by $A_\theta(\cdot)$, and augmented input by
\begin{equation}
x' := A_\theta(x).
\label{eq:xprime_def}
\end{equation}
To analyze its effect, we decompose the augmentation into a structured transformation term and an additive localized perturbation term:
\begin{equation}
A_\theta(x) = T_\theta(x) + \Delta_\theta(x).
\label{eq:aug_decomp}
\end{equation}
Here, $T_\theta(x)\in[0,1]^{C\times H\times W}$ denotes the structured transform component, and $\Delta_\theta(x)\in\mathbb{R}^{C\times H\times W}$ denotes the additive perturbation component. In our setting, the structured transform is saliency-guided and corresponds to the selective rotation/blur step (Eq.~\ref{equ:three}) aggregated over the accepted patches (Algorithm~\ref{alg:multiscale_mixup}). Let $S(x)\in[0,1]^{1\times H\times W}$ be a per-pixel saliency map, and let $M_\theta\in\{0,1\}^{1\times H\times W}$ be an effective binary mask induced by the accepted patch-level saliency masks after resizing to full resolution. Let $R(x,\varphi)$ denote a rotation of $x$ by angle $\varphi$, where $\varphi$ is sampled as part of $\theta$, and let $B(\cdot)$ denote Gaussian blurring. With the same mixing coefficient $\alpha\sim\text{Uniform}(0,1)$ as in Eq.~(\ref{euq:equ8}), we define
\begin{equation}
T_\theta(x)=\alpha x+(1-\alpha)\Big(R(x,\varphi)\odot M_\theta + B(x)\odot (1-M_\theta)\Big),
\label{eq:Ttheta}
\end{equation}
where $\odot$ denotes element-wise multiplication. We further model the localized fractal perturbation consistent with Sec.~\ref{sub:fractal_minmodes} fractals are injected only through the selected region, using the same blending factor $\lambda$ as Eq.~(\ref{equ:six}):
\begin{equation}
\Delta_\theta(x)=(1-\alpha)\lambda\,(M_\theta\odot F),
\label{eq:Deltatheta}
\end{equation}
where $F\in\mathbb{R}^{C\times H\times W}$ denotes the (full-resolution) fractal pattern corresponding to the sampled fractal(s) after resizing/aggregation. Then, we analyze the augmentation under the vicinal risk minimization (VRM) \cite{chapelle2000vicinal} objective:
\begin{equation}
\mathrm{VRM}(h_\omega):=\mathbb{E}_{(x,y)\sim\mathcal{D}}\;\mathbb{E}_{\theta\sim\Pi}\big[\ell(h_\omega(A_\theta(x)),y)\big]
=\mathbb{E}_{x,y}\mathbb{E}_{\theta}\big[g(T_\theta(x)+\Delta_\theta(x))\big].
\label{eq:vrm_def}
\end{equation}
For brevity, let
\begin{equation}
x_T := T_\theta(x), \qquad \Delta_\theta := \Delta_\theta(x),
\vspace{-5pt}
\end{equation}
so that $x'=x_T+\Delta_\theta$. We make two assumptions. The fractal perturbation is zero-mean with covariance $\Sigma_f$, i.e., $\mathbb{E}[F]=0$ and $\mathrm{Cov}(F)=\Sigma_f$. Since $\Delta_\theta=(1-\alpha)\lambda(M_\theta\odot F)$ and $\alpha$ is sampled independently, this implies $\mathbb{E}_\theta[\Delta_\theta]=0$. Label consistency under the structured transform $T_\theta$: applying $T_\theta$ does not alter the semantic label $y$. This assumption is used only to justify evaluating the same target $y$ after the transformation. We now perform a second-order Taylor expansion of $g$ around $x_T$. Since $x'=x_T+\Delta_\theta$, we have
\begin{equation}
g(x_T+\Delta_\theta)
\approx
g(x_T)+\nabla g(x_T)^\top \Delta_\theta+\frac{1}{2}\Delta_\theta^\top H_g(x_T)\Delta_\theta,
\label{eq:taylor_expansion}
\end{equation}
where $\nabla g(x_T)$ and $H_g(x_T)$ denote the gradient and Hessian of $g$ with respect to the input, evaluated at $x_T$. Taking expectation over $\theta$, the first-order term vanishes due to the zero-mean perturbation assumption:
\begin{equation}
\mathbb{E}_\theta[\nabla g(x_T)^\top \Delta_\theta]
=
\nabla g(x_T)^\top \mathbb{E}_\theta[\Delta_\theta]
=0.
\label{eq:first_order_vanish}
\end{equation}
Therefore, the expected loss under augmentation is approximately
\begin{equation}
\mathbb{E}_\theta[g(x_T+\Delta_\theta)]
\approx
\mathbb{E}_\theta[g(x_T)]
+\frac{1}{2}\mathbb{E}_\theta\!\left[\Delta_\theta^\top H_g(x_T)\Delta_\theta\right].
\label{eq:expected_taylor}
\end{equation}
Taking expectation over $(x,y)\sim\mathcal{D}$ yields
\begin{equation}
\mathrm{VRM}(h_\omega)
\approx
\mathbb{E}_{x,y}\mathbb{E}_\theta[g(x_T)]
+\frac{1}{2}\mathbb{E}_{x,y}\mathbb{E}_\theta\!\left[\Delta_\theta^\top H_g(x_T)\Delta_\theta\right].
\label{eq:vrm_after_taylor}
\end{equation}
To interpret the second term, we rewrite the quadratic form in trace form using the identity $\Delta^\top H\Delta=\mathrm{tr}(H\Delta\Delta^\top)$. This gives
\begin{equation}
\mathbb{E}_\theta\!\left[\Delta_\theta^\top H_g(x_T)\Delta_\theta\right]
=
\mathrm{tr}\!\left(H_g(x_T)\,\mathbb{E}_\theta[\Delta_\theta\Delta_\theta^\top]\right).
\label{eq:trace_rewrite}
\end{equation}
We now compute second moment of the perturbation. Since $\Delta_\theta=(1-\alpha)\lambda(M_\theta\odot F)$ and $\alpha\sim\text{Uniform}(0,1)$ is independent of $(M_\theta,F)$, we have $\mathbb{E}[(1-\alpha)^2]=\frac{1}{3}$ and thus
\begin{equation}
\mathbb{E}_\theta[\Delta_\theta\Delta_\theta^\top]
=
\frac{\lambda^2}{3}\,\mathbb{E}_\theta\!\left[M_\theta\,\Sigma_f\,M_\theta^\top\right].
\label{eq:delta_cov}
\end{equation}
We define mask-dependent covariance as localized perturbation covariance:
\begin{equation}
\Sigma_{\mathrm{loc}}(x):=\mathbb{E}_\theta\!\left[M_\theta\,\Sigma_f\,M_\theta^\top\right].
\label{eq:sigma_loc}
\end{equation}
Substituting Eq.~\eqref{eq:delta_cov} and Eq.~\eqref{eq:sigma_loc} into Eq.~\eqref{eq:trace_rewrite} yields
\begin{equation}
\mathbb{E}_\theta\!\left[\Delta_\theta^\top H_g(x_T)\Delta_\theta\right]
=
\frac{\lambda^2}{3}\,\mathrm{tr}\!\left(H_g(x_T)\Sigma_{\mathrm{loc}}(x)\right).
\label{eq:quadratic_final}
\end{equation}
Hence, the vicinal risk admits the following second-order approximation:
\begin{equation}
\mathrm{VRM}(h_\omega)
\approx
\underbrace{\mathbb{E}_{x,y}\mathbb{E}_\theta\big[\ell(h_\omega(T_\theta(x)),y)\big]}_{\text{invariance term}}
+
\underbrace{\frac{\lambda^2}{6}\,\mathbb{E}_{x}\!\left[\mathrm{tr}\!\left(H_g(T_\theta(x))\Sigma_{\mathrm{loc}}(x)\right)\right]}_{\text{saliency-local stability penalty}}.
\label{eq:vrm_decomposition_final}
\end{equation}
The first term enforces invariance to the structured saliency-guided transform $T_\theta$, while the second penalizes curvature-weighted sensitivity to fractal perturbations restricted by $\Sigma_{\mathrm{loc}}(x)$, concentrating the regularization on accepted salient regions (Algorithm~\ref{alg:multiscale_mixup}). This characterization yields two predictions borne out by our experiments. The penalty grows quadratically in $\lambda$, so accuracy should first improve and then degrade as $\lambda$ increases, matching Figure~\ref{fig:hyparameter}(b) where performance peaks at $\lambda$=0.20 and drops at $\lambda$=0.50. Moreover, since the penalty suppresses sensitivity to localized input perturbations, the largest gains should appear under corruption and adversarial noise rather than on clean data, matching Table~\ref{tab:4} where our corruption gain (\gain{2.40\%}) and FGSM error reduction (\gain{3.14\%}) over AdAutoMix both exceed the clean gain (\gain{1.19\%}). The choice between local and global injection corresponds to different $\Sigma_{\mathrm{loc}}(x)$, and we resolve that comparison empirically in Table~\ref{tab:tab15}.

\begin{table}[b]
    \centering
        \vspace{-5pt}
    \setlength{\tabcolsep}{1.5mm}

    \caption{Top-1 performance (\%)$\uparrow$ of mixup methods on CIFAR-100, Tiny-ImageNet and ImageNet-1K. The results of previous mixup SOTA methods are taken from AdAutoMix \cite{qinadversarial}. Res18, ResXt50 CNext-T and Res34 refers to ResNet18, ResNeXt50, ConvNeXt-T and ResNet34. Also, ViT-B results are taken from \cite{bai2022improving}.}
    \vspace{-10pt}

    \resizebox{1.0\linewidth}{!}{
    \begin{tabular}{l|cc|cc|cc|ccc|c} \toprule
            & \multicolumn{2}{c|}{\textbf{CIFAR-100}}    
            & \multicolumn{2}{c|}{\textbf{CIFAR-100}} 
            & \multicolumn{2}{c|}{\textbf{Tiny-ImageNet}}  
            & \multicolumn{4}{c}{\textbf{ImageNet-1K}} \\
    \textbf{Method} 
            & \textbf{Res18}       
            & \textbf{ResXt50}          
            & \textbf{Swin-T}          
            & \textbf{CNeXt-T}   
            & \textbf{Res18}       
            & \textbf{ResXt50}  
            & \textbf{Res18}    
            & \textbf{Res34}    
            & \textbf{Res50} 
            & \textbf{ViT-B}  \\ \midrule
    Vanilla         
            & 78.04 & 81.09 & 78.41 & 78.70 & 61.68 & 65.04 
            & 70.04 & 73.85 & 76.83 & 76.7 \\
    MixUp           
            & 79.12 & 82.10 & 76.78 & 81.13 & 63.86 & 66.36 
            & 69.98 & 73.97 & 77.12 & 80.8 \\
    CutMix          
            & 78.17 & 81.67 & 80.64 & 82.46 & 65.53 & 66.47 
            & 68.95 & 73.58 & 77.17 & 79.9 \\
    SaliencyMix     
            & 79.12 & 81.53 & 80.40 & 82.82 & 64.60 & 66.55 
            & 69.16 & 73.56 & 77.14 & -- \\
    FMix            
            & 79.69 & 81.90 & 80.72 & 81.79 & 63.47 & 65.08 
            & 69.96 & 74.08 & 77.19 & -- \\
    PuzzleMix       
            & 81.13 & 82.85 & 80.33 & 82.29 & 65.81 & 67.83 
            & 70.12 & 74.26 & 77.54 & -- \\
    ResizeMix       
            & 80.01 & 81.82 & 80.16 & 82.53 & 63.74 & 65.87 
            & 69.50 & 73.88 & 77.42 & -- \\
    AutoMix         
            & 82.04 & 83.64 & 82.67 & 83.30 & 67.33 & 70.72 
            & 70.50 & 74.52 & 77.91 & -- \\
    AdAutoMix       
            & 82.32 & 84.22 & 84.33 & 83.54 & 69.19 & 72.89 
            & 70.86 & 74.82 & 78.04 & -- \\ \midrule 
    \bf{\our{}}     
            & \bf{82.74} & \bf{84.91} & \bf{85.35} & \bf{84.41} 
            & \bf{70.38} & \bf{74.27} 
            & \bf{71.37} & \bf{75.34} & \bf{78.54} & \textbf{ 81.2} \\ 
    \bottomrule
    \end{tabular}
    }
    \vspace{2pt}

    \label{tab:coarse_grained}
    \vspace{-10pt}
\end{table}

\section{Experiments and Results}
We benchmark our proposed \our{} against several recent competitive mixup approaches, including Mixup~\cite{zhang2018mixup}, CutMix~\cite{yun2019cutmix}, ManifoldMix~\cite{verma2019manifold}, FMix~\cite{harris2020fmix}, ResizeMix~\cite{qin2020resizemix}, SaliencyMix~\cite{uddin2020saliencymix}, PuzzleMix~\cite{kim2020puzzle},  AutoMix~\cite{liu2022automix}, and AdAutoMix~\cite{qinadversarial}. Additionally, we report the computational overhead of our method and compare it with timings reported in AdAutoMix~\cite{kang2023guidedmixup}. To demonstrate the generalizability of our method, we conduct experiments from small-scale to large-scale backbones including \texttt{ResNet18} \cite{he2016deep}, \texttt{ResNet34} \cite{he2016deep}, \texttt{ResNet50}~\cite{he2016deep}, \texttt{ResNeXt50}~\cite{xie2017aggregated}, transformer-based architectures including \texttt{Swin Transformer}~\cite{liu2021swin} and \texttt{ConvNeXt}~\cite{liu2022convnet}, and contrastive method \texttt{MoCo v2} \cite{chen2020improved} and \texttt{SimSiam} \cite{chen2021exploring}. All experiments are implemented using open-source OpenMixup.
\par
Note that, we follow the standard evaluation practices and protocols mentioned in AdAutoMix \cite{qinadversarial} for fair comparison. Hyperparameter configurations and brief \textit{implementation guidelines}, with  \textit{dataset statistics} are also provided in \textcolor{magenta}{Appendix A} and \textit{detailed settings} provided in \textcolor{magenta}{Appendix B}. We show that our proposed \our{} not only improves classification performance across both general- and fine-grained tasks, but also enhances robustness to distributional shifts, such as background corruption~\cite{hendrycks2020augmix}, data scarcity, transfer learning, calibration, contrastive learning methods, object detection while maintaining minimal computational overhead. 


\subsection{General Classification}
We compare the performance of \our{} in Tab. \ref{tab:coarse_grained}, our approach achieves SOTA performance on CNNs and ViTs, consistently outperforming existing augmentation strategies such as AdAutoMix \cite{qinadversarial}, AutoMix, ResizeMix, and PuzzleMix. Notably, \our{} surpasses AdAutoMix, the existing best performing method \cite{qinadversarial}, by approximately \gain{0.42\%} and \gain{0.69\%} in Top-1 accuracy on CIFAR-100. The trend is similar across different backbones from small-scale to large-scale backbones. In terms of Tiny-ImageNet and ImageNet-1K, the improvement gap is even more pronounced, underlining \our{} capacity to capture rich discriminative features. These results demonstrate that \our{} not only addresses the challenges inherent in CIFAR-100, Tiny-ImageNet, and ImageNet-1K but also establishes a new SOTA among mixup based augmentation methods for enhancing generalization performance.

\begin{table}[t]
    \centering
    \setlength{\tabcolsep}{2.0mm}
    \caption{Accuracy (\%) $\uparrow$ of mixup methods on Caltech Birds-200, FGVC-Aircrafts and Stanford-Cars.}

    \vspace{-10pt}
\resizebox{0.85\linewidth}{!}{
    \begin{tabular}{l|cc|cc|cc} \toprule
                    & \multicolumn{2}{c|}{ \textbf{Caltech Birds-200} }      & \multicolumn{2}{c|}{ \textbf{FGVC-Aircrafts}}
                    & \multicolumn{2}{c}{ \textbf{Stanford-Cars}}\\
  \textbf{ Method}           & \textbf{ResNet18}        & \textbf{ ResNet50}     & \textbf{ResNet18}       & \textbf{ResNeXt50}        & 
\textbf{ResNet18}       & \textbf{ResNeXt50}   \\ \midrule
    Vanilla         & 77.68         & 82.38        & 80.23         & 85.10           & 86.32         & 90.15      \\
    MixUp           & 78.39         & 82.98        & 79.52         & 85.18           & 86.27         & 90.81      \\
    CutMix          & 78.40         & 83.17        & 78.84         & 84.55           & 87.48         & 91.22      \\
    ManifoldMix     & 79.76         & 83.76        & 80.68         & 86.60           & 85.88         & 90.20      \\
    SaliencyMix     & 77.95         & 81.71        & 80.02         & 84.31           & 86.48         & 90.60      \\
    FMix            & 77.28         & 83.34        & 79.36         & 86.23           & 87.55         & 90.90   \\
    PuzzleMix       & 78.63         & 83.83        & 80.76         & 86.23           & 87.78         & 91.29     \\
    ResizeMix       & 78.50         & 83.41        & 78.10         & 84.08           & 88.17         & 91.36      \\
    AutoMix         & 79.87   &  83.88  &  81.37   &  86.72     & 88.89   & 91.38 \\ 
    AdAutoMix  &  80.88     &  84.57    &  81.73     & 87.16       & 89.19     & 91.59 \\ \midrule 
    \textbf{\our{}}  & \textbf{81.84}       & \textbf{85.73}      & \textbf{82.81}      & \textbf{88.34}        & \textbf{90.56}      & \textbf{92.86} \\
    \bottomrule
    \end{tabular}
    }

    \label{tab:fgvc_acc}

\end{table}

\begin{table}[t]
    \centering
        \caption{Top-1 accuracy (\%) of ResNet-50 (R50) and Vision Transformer (ViT) backbones on CUB-200 and Stanford Cars datasets.}

    \vspace{-10pt}
    \resizebox{0.95\linewidth}{!}{
    \begin{tabular}{c|l|cccccc|c} \toprule
    \textbf{Backbone}  & \textbf{ Dataset} & \textbf{ Vanilla} & \textbf{MixUp}  & \textbf{ CutMix} & \textbf{PuzzleMix}  & \textbf{ AutoMix} & \textbf{AdAutoMix}  & \textbf{\our{}} \\ \midrule
    
    \multirow{2}{*}{\rotatebox{90}{R50}} 
        & CUB-200       & 81.76 & 82.79 & 81.67 & 82.59 & 82.93 & 83.36 & \textbf{84.42} \\
        & Stanford Cars & 88.88 & 89.45 & 88.99 & 89.37 & 88.71 & 89.65 & \textbf{90.85} \\ \midrule
    
    \multirow{2}{*}{\rotatebox{90}{ViT-B}} 
        & CUB-200       & 88.0 & 88.75 & 87.76 & 88.23 & 88.91 & 88.76 & \textbf{89.84} \\
        & Stanford Cars & 91.31 & 91.36 & 91.53 & 91.83 & 92.51 & 91.38 & \textbf{92.86} \\ \bottomrule
    
    \end{tabular}}


        \label{lab:transfrer_learning}

\end{table}

\vspace{-3pt}
\subsection{Fine-Grained Visual Classification} 
\vspace{-3pt}
In fine-grained classification, we follow the same training protocols established in AdAutoMix~\cite{qinadversarial} and also previous results are taken from same paper.
As reported in Tab.~\ref{tab:fgvc_acc}, \our{} consistently achieves the highest Top-1 accuracy across different architectures and datasets. On Caltech Birds-200, \our{} improves over the AdAutoMix by \gain{+0.96\%} on \texttt{ResNet-18} and \gain{+1.16\%} on \texttt{ResNet-50}. On FGVC-Aircrafts, it achieves gains of \gain{+1.08\%} and \gain{+1.18\%} on \texttt{ResNet-18} and \texttt{ResNeXt-50}, respectively. On Stanford-Cars, improvements of \gain{+1.37\%} and \gain{+1.27\%} are observed with \texttt{ResNet-18} and \texttt{ResNeXt-50}. These results establish the  effectiveness of \our{} across fine-grained  categorization tasks.

\subsection{Transfer Learning}
We further evaluate the transferability of features learned by \our{} on downstream transfer learning classification tasks, as presented in Table~\ref{lab:transfrer_learning}. We utilized two  pre-trained deep models including  \texttt{ResNet-50} and \texttt{ViT-B}. Both models are pretrained on ImageNet-1K and fine-tuned on Caltech Birds-200 and Stanford-Cars for classification using \our{}. Compared to AdAutoMix ~\cite{qinadversarial}, the strongest existing method, \our{} achieves consistent gains. In Table \ref{lab:transfrer_learning} on Caltech Birds-200, \our{} reaches a Top-1 performances of \gain{84.42\%, 89.84\%}, outperforming the AdAutoMix by \gain{1.06\%, 1.08\%}. On the Stanford-Cars, it achieves \gain{90.85\%, 92.86\%}, exceeding the baseline by \gain{1.20\%, 1.48\%}. These results demonstrate that \our{} improves fine-tuning performance over baseline and recent SOTA methods across different datasets.

\subsection{Robustness}
\begin{wraptable}{r}{0.49\linewidth} 
  \centering
  \vspace{-40pt} 
      \caption{Top-1 accuracy and FGSM error of \texttt{ResNet-18} with other methods.}

\resizebox{\linewidth}{!}{
\begin{tabular}{l|ccc}
\toprule
                & \textbf{Clean}              & \textbf{Corruption}          & \textbf{FGSM}  \\
\textbf{Method}          &  \textbf{Acc(\%)$\uparrow$} & \textbf{Acc(\%)$\uparrow$}  & \textbf{ Error(\%)$\downarrow$} \\
\midrule
CutMix          & 79.45             & 46.66              & 88.24 \\
FMix            & 78.91             & 50.58              & 88.35 \\
PuzzleMix       & 79.96             & 51.04              & 80.52 \\
AutoMix         & 80.02             & 50.75              & 82.67 \\
AdAutoMix       & 81.55             & 51.44              & 75.66 \\
\midrule
\textbf{\our{}} & \textbf{82.74}    & \textbf{53.84}     & \textbf{72.52} \\
\bottomrule
\end{tabular}
}


      \label{tab:4}
\end{wraptable}
Following the same protocols as used by AdAutoMix \cite{qinadversarial}, we carried out robustness evaluation experiments under common corruptions on CIFAR100-C \cite{hendrycks2019robustness} dataset as shown in Tab. \ref{tab:4}. We compared our \our{} with widely used mixup approaches, including CutMix, FMix, PuzzleMix, AutoMix, and AdAutoMix \cite{qinadversarial}.  As shown in Tab. \ref{tab:4}, \our{} demonstrated the best performance on both clean and corrupted samples, achieving relative gains of \gain{1.19\%} and \gain{2.4\%} in classification accuracy over AdAutoMix. The robustness improvement of \gain{3.14\%} is achieved compared to AdAutoMix.
\begin{figure*}[t]
    \centering
    \includegraphics[width=1.0\textwidth]{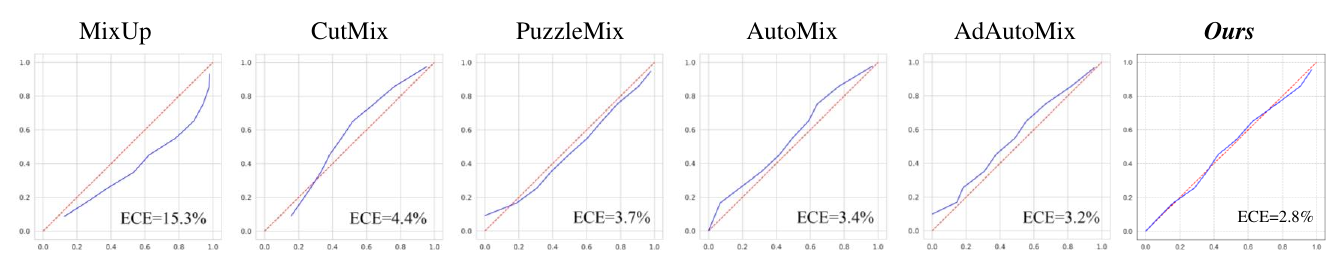}
    \caption{Calibration plots of \our{} on CIFAR-100 using \texttt{ResNet18}. 
    }
    \label{fig:calibration}
\end{figure*}
\subsection{Calibration}
\label{sec:calibration}
Deep Neural Networks often exhibit overconfidence in their predictions during image classification tasks, which can lead to poor calibration. To quantitatively assess calibration performance, we measure the Expected Calibration Error (ECE) across different mixup methods on the CIFAR-100 dataset. Previous results are taken from AdAutoMix \cite{qinadversarial}. As visbible in Fig. \ref{fig:calibration}, our  \our{} attains the lowest ECE \gain{2.8\%} surpassing recent SOTA methods and second best method is AdAutoMix \cite{qinadversarial}. 


\begin{table}[t]
\centering

\begin{minipage}[t]{0.47\linewidth}
\vspace{15pt}
\centering
\includegraphics[width=\linewidth]{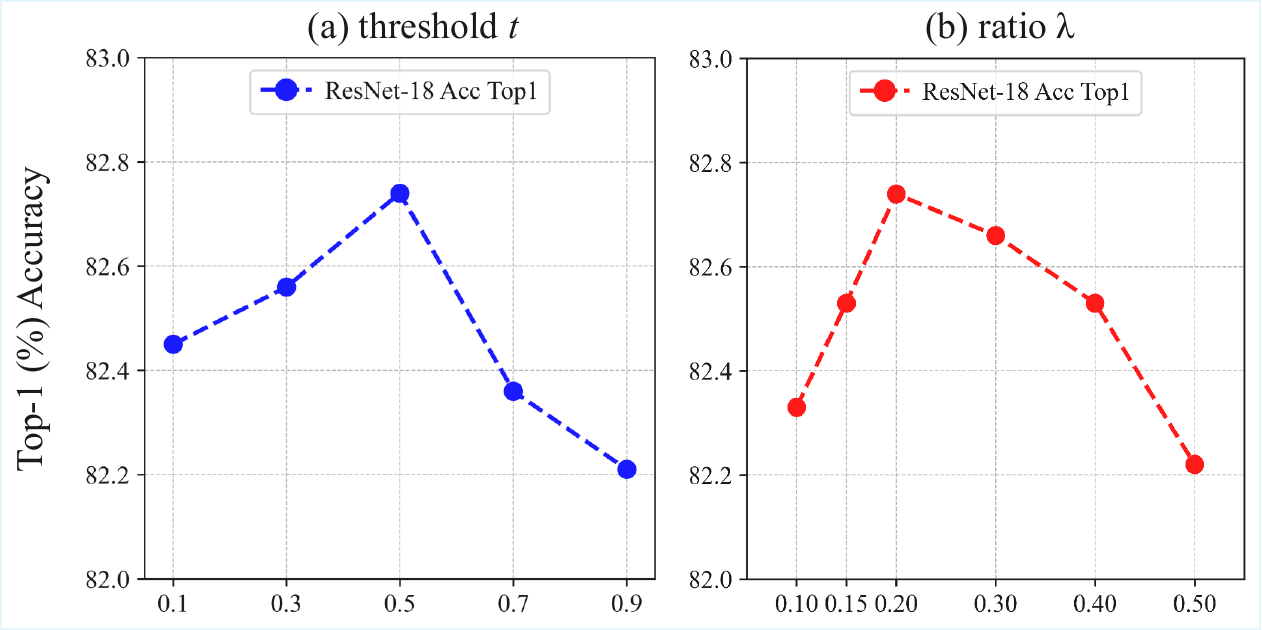}
\vspace{2pt}
\captionof{figure}{Ablation of hyperparameters $t$ threshold and $\lambda$ for fractal mixing of \our{} on CIFAR100.}
\label{fig:hyparameter}
\vspace{-10pt}
\end{minipage}\hspace{10pt}%
\begin{minipage}[t]{\dimexpr\linewidth-0.47\linewidth-10pt\relax}
\vspace{0pt}
\centering
\captionof{table}{Ablation study of different high-level mixing strategies on CIFAR-100.} 
\vspace{-10pt}
\setlength{\tabcolsep}{1.0mm}
\scalebox{0.80}{
\begin{tabular}{c c c c c c}
\toprule
\multicolumn{4}{c}{\textbf{CIFAR-100}} & \multicolumn{2}{c}{\textbf{Accuracy (\%)}} \\
\cmidrule(lr){1-4} \cmidrule(lr){5-6}
\textbf{\our{}} & \textbf{$M_m$} & \textbf{$M_c$} & \textbf{$M_r$} & \textbf{Res18} & \textbf{ResXt50} \\
\midrule
-- & -- & -- & -- & 78.04 & 81.09 \\
\checkmark & -- & -- & -- & 81.73 & 82.22 \\
-- & \checkmark & -- & -- & 79.12 & 82.10 \\
-- & -- & \checkmark & -- & 78.17 & 81.67 \\
-- & -- & -- & \checkmark & 80.01 & 81.82 \\
\checkmark & \checkmark & \checkmark & -- & 82.24 & 82.32 \\
\checkmark & \checkmark & -- & \checkmark & 82.46 & 82.89 \\
\checkmark & -- & \checkmark & \checkmark & 82.58 & 83.52 \\
\checkmark & \checkmark & \checkmark & \checkmark & \textbf{82.74} & \textbf{84.91} \\
$M_f$ & \checkmark & \checkmark & \checkmark & 80.24 & 82.27 \\
\bottomrule
\end{tabular}
}
\vspace{2pt}

\label{tab:ablation}

\end{minipage}
\end{table}

\section{Ablation Study and Analysis}
\label{sec:Ablation}
As noted in Sec.~\ref{sec:PM}, our technique allows multiple modes. Here, we analyze the effects of those modes on performance. More results are mentioned in \textcolor{magenta}{Appendix~C}.  

\vspace{1mm}
\noindent{\textbf{Inclusion of Simple Modes.}} 
Tab. \ref{tab:ablation} presents  \texttt{Res18} and \texttt{ResXt50} results on CIFAR-100. In the table, $M_m$ denotes Mixup \cite{guo2019mixup}, $M_c$ represents CutMix \cite{yun2019cutmix}, $M_r$ presents ResizeMix \cite{qin2020resizemix} and $M_f$ illustrates FMix\cite{harris2020fmix}. We start with our \our{}, which offers two main improvements: i) it generates multi-scale features. ii) saliency-driven patch transformations in a more principled and diverse manner. The introduction of the \our{} leads to a significant gain of 3.69\% and 1.13\% in terms of performance over the base model, highlighting the impact of self-mixing compared to the individual performance of other modes $M_m$, $M_c$ and $M_r$.  To enable comparison  to  FMix \cite{harris2020fmix}, while retaining the same training and implementation details, we use the mixing mode ``$M_f$+$M_m$+$M_c$+$M_r$". Clearly, this  degrades the overall performance, which highlights the efficacy of  \our{} in the mixed mode. 

\vspace{5pt}
\noindent{\textbf{Exclusion of Simple Modes.}} In our proposed high-level mixing we do not select methods such as PuzzleMix~\cite{kim2020puzzle}, Co-Mixup~\cite{kim2021co}, and GuidedMixup~\cite{kang2023guidedmixup}, which  demonstrate good performance but incur high computational overhead. As mentioned in Tab. \ref{tab:ablation}, the best combination is ``\our{}+$M_m$+$M_c$+$M_r$" and the reason  is that $M_m$ complements $S^2$ in terms of global inter-image variations, while $M_c$ introduces local inter-image diversity. $M_r$ introduces down-scaled inter-image variations which are missing in the other modes. Thus, the selected set of modes complement each other well to generate a diverse set of augmentations. 

\vspace{5pt}
\noindent{\textbf{Motivation behind High-level Mixing.}}
Four crucial objectives of the current augmentation methods include: \textit{scale-invariance}, \textit{inter-image diversity}, \textit{spatial variability}, and \textit{resolution robustness}. Previous methods address these challenges in isolation. With high-level mixing, our  objective is to jointly tackles all four objectives  while maintaining low computational overhead. Replacing \our{} with $M_f$ (as used by \cite{liu2025randomix}) while keeping all remaining settings same in Tab.~\ref{tab:ablation}, leads to significant performance reduction. This verifies that our high-level mixing achieves the intended objective through correct composition of modes. 

\begin{table}[t]
\centering
\vspace{-4pt}

\begin{minipage}[t]{0.49\linewidth}
\centering
\captionof{table}{Ablation study of key components of FracMix in terms of global and local fractals on Stanford-Cars. $sal$ denotes saliency and $weight$ denotes weighting.}
\vspace{-10pt}
\scalebox{0.90}{
\begin{tabular}{lc}
\toprule
\textbf{Method}  & \textbf{Top-1 (\%)}  \\
\midrule
Baseline                         & 85.52 \\
\our{} ($w/o$ $sal$ $weight$)    & 91.87 \\
\our{} ($global$ $fractal$)      & 92.27 \\
\textbf{\our{} ($local$)}        & \textbf{92.78} \\
\bottomrule
\end{tabular}
}

\label{tab:tab15}
\end{minipage}
\hfill
\begin{minipage}[t]{0.49\linewidth}
\centering
\captionof{table}{Top-1 (\%) performance ablation of saliency vs.\ fractal mixing on CIFAR-100 with ResNet-18 and ResNeXt-50.}
\vspace{-10pt}
\setlength{\tabcolsep}{5.5pt}
\scalebox{0.85}{
\begin{tabular}{lcc}
\toprule
\textbf{Method} &  \textbf{Res18} & \textbf{ResXt50} \\
\midrule
Baseline                 & 78.04 & 81.09 \\
Saliency (A+B)           & 79.12 & 81.53 \\
Saliency S$^{2}$ (A+A)   & 79.54 & 81.92 \\
\midrule
FracMix (ours)           & 81.73 & 82.22 \\
\textbf{\our{}}          & \textbf{82.74} & \textbf{84.91} \\
\bottomrule
\end{tabular}
}

\label{tab:resnet_resnext_fracmix}
\end{minipage}

\end{table}

\vspace{5pt}
\noindent{\textbf{Global Fractal vs Local Fractal.}}
In Tab.~\ref{tab:tab15}, we analyze design choices for fractal addition on Stanford-Cars. We compare \our{} with saliency-weighted fractal addition applied only to salient regions (Eq. \ref{equ:six}) against a global variant that adds fractals to the entire image, $I_i=\lambda F + (1-\lambda)I_i$. We further study \our{} without saliency weighting by modifying (Eq. \ref{equ:three}) to use an equal mixture of region and background transformations,
$T_k = 0.50 \, R(P^f_k, \theta) + 0.50 \, B(P^f_k)$.
These comparisons isolate the benefit of focusing fractal perturbations on salient content rather than applying them uniformly.

\vspace{5pt}
\noindent{\textbf{Effect of Saliency and Fractal Mixing.}} We also study how the mixing strategy influences CIFAR-100 accuracy across Res18 and ResXt-50 backbones. We compare a baseline with saliency-guided mixing, where Saliency (A+B) uses the saliency map from the source image $A$ to guide mixing with a target image $B$, and Saliency S$^{2}$ (A+A) uses self-saliency from both inputs presented in Tab.~\ref{tab:resnet_resnext_fracmix}. While saliency improves over the baseline on both backbones, our fractal-based \textit{FracMix} yields a larger gain, and the full method \our{} achieves the best performance overall. We attribute this synergy to the fact that salient regions harbor the most discriminative semantic features; by injecting self-similar fractal patterns specifically into these regions, we force the model to learn robust features invariant to high-frequency perturbations where it matters most, rather than diluting regularization energy on non-discriminative background pixels.

\vspace{5pt}
\noindent{\textbf{Hyperparameters Ablation.}}
\label{main_paper:Hyperparameters}
In the \our{}, there are two main hyperparameters namely saliency threshold $t$ and $\lambda$. In order to achieve good performance both parameters should be properly configured. Firstly, we train the  \texttt{ResNet18} for 200 epochs via our \our{}. The performance of \texttt{ResNet18} with $t$=$0.5$ is shown in Fig. \ref{fig:hyparameter} (a). In addition, the fractal mixing gives best performance at $\lambda$ = $0.20$ in Fig.  \ref{fig:hyparameter} (b). However, by increasing the $\lambda$ to $0.50$, we observe that the classification accuracy is slightly decreased to 82.22\%, and when we set $t$=$0.9$ the performance degrades to almost the same level. This implies that the two parameters are equally capable of controlling the performance of \our{}.

\begin{figure}[t]
        \centering
        \includegraphics[width=0.90\linewidth]{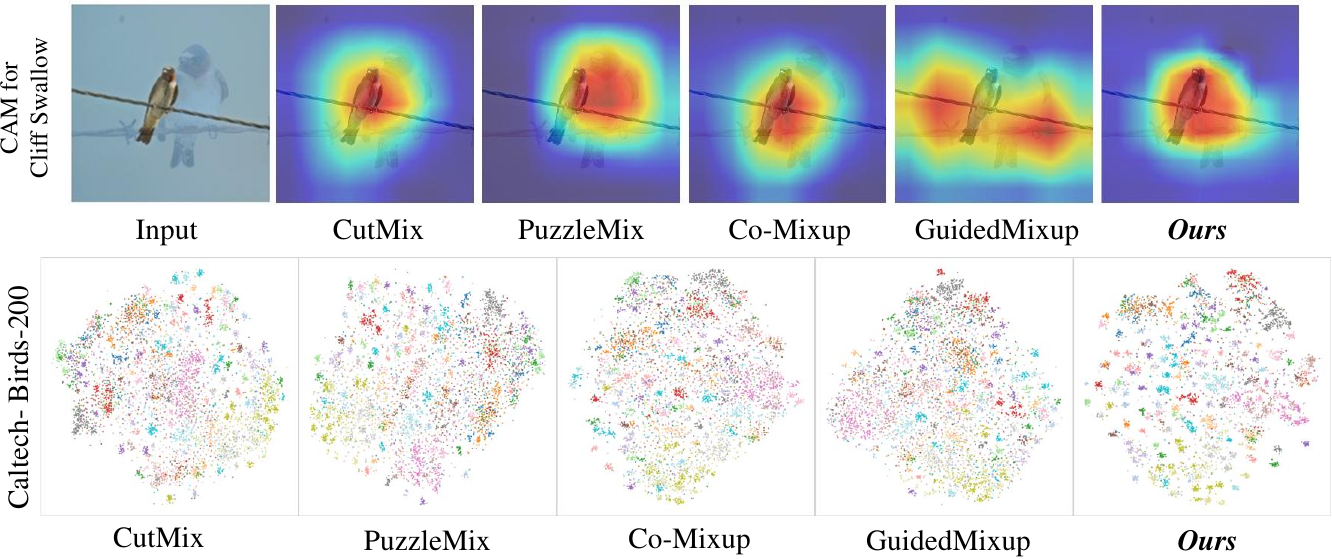}
        \caption{(Top) Grad-CAM \cite{selvaraju2020grad} visualization on augmented images and (bottom) t-SNE visualization of \texttt{ResNet18} trained from scratch. 
        }
        \vspace{-18pt}
        \label{fig:gradcam_main}

\end{figure}

\vspace{5pt}
\noindent\textbf{Object Localization.} For this analysis, we visually analyze the models trained with \our{} and SOTA methods. As evidenced in Fig. \ref{fig:gradcam_main} (top row), our proposed \our{} produces a contiguous, high-intensity CAM region that consistently highlights the main region, indicating stronger object retention and clearer attention boundaries. 

\vspace{5pt}
\noindent{\textbf{Feature Representation.}} Finally, we compare the trained models by visualizing the feature representation of \our{} and SOTA methods in Fig. \ref{fig:gradcam_main} (bottom row). Closely observing t-SNE \cite{van2008visualizing}, it can be seen that images of the same class cluster together for our method, representing better learning. Noticeably, \our{} exhibits distinct clusters with well-defined margins between classes, suggesting that the model consistently learns discriminative features specific to each class.

\section{Conclusion}

We introduce \our{} data augmentation technique to improve the performance of deep visual models. The method comprises two main components. In the $S^2$ mixing component, patches of varying sizes are extracted from an input image while utilizing  saliency information. Different transformations are applied to these patches and they are seamlessly integrated back into the same image. In the  \textit{FracMix} component, self-similarity fractals are also blended into these salient patches. We also introduce the notion of  high-level mixing of multiple low-level mixing modes to enhance diversity among the augmented samples. Experiments are performed on coarse and fine-grained classification, robustness against corruption, few-shot learning, and transfer learning. The proposed \our{} has demonstrated improved results compared to the existing state-of-the-art methods. The potential {limitations}  of our approach are also discussed in  \textcolor{magenta}{Appendix~D}.

\section{Acknowledgement}
Naveed Akhtar is a recipient of the Australian Research Council Discovery Early Career Researcher Award (project \# DE230101058) funded by the Australian Government. Khawar Islam is also supported by project \# DE230101058. This research was also supported by The University of Melbourne’s Research Computing Services and the Petascale Campus Initiative.

\bibliographystyle{splncs04}
\bibliography{main}

\clearpage
\appendix

\section*{Overview of the Appendices}
\addcontentsline{toc}{section}{Overview of the Appendices}

This supplementary material complements the main paper with the following content.

\begin{itemize}
    \item \textbf{Section A.} Details of the datasets used for training and testing.
    \item \textbf{Section B.} Experimental settings, hyperparameters configurations.
    \item \textbf{Section C.} Additional ablation studies of our proposed method.
    \item \textbf{Section D.} Limitations of our proposed \our{} method.
\end{itemize}

\section{Dataset Information}
\label{sec:data_info}
We provide a concise overview of the datasets used in our study, following the protocols adopted by prior SOTA mixup methods~\cite{guo2019mixup,qinadversarial,yun2019cutmix,islam2024diffusemix}.

\vspace{10pt}
\noindent \textbf{CIFAR-100}~\cite{Krizhevsky09learningmultiple} is a widely used benchmark dataset for evaluating computer vision models. It comprises $50{,}000$ training images and $10{,}000$ test images, each with a resolution of $32 \times 32$. The images are evenly distributed across $100$ fine-grained classes, with each class containing $600$ images ($500$ for training and $100$ for testing).

\vspace{10pt}
\noindent \textbf{Tiny-ImageNet}~\cite{chrabaszcz2017downsampled} is a compact version of the ImageNet dataset~\cite{imagenet}. It consists of $200$ classes, each containing $500$ training images, $50$ validation images, and $50$ test images. Each image is resized to a fixed resolution of $64 \times 64$ pixels, making the dataset computationally efficient for training and evaluation.

\vspace{10pt}
\noindent \textbf{ImageNet-1K}~\cite{imagenet} is a large-scale image classification benchmark consisting of approximately $1.28$ million training images distributed across $1{,}000$ classes. Each class contains around $1{,}300$ images, and the dataset also includes a validation set of $50{,}000$ images. The image resolutions are variable but generally exceed $256\times256$ pixels, with resizing applied during preprocessing.

\vspace{10pt}
\noindent \textbf{CUB-200-2011}~\cite{WahCUB_200_2011} is a fine-grained dataset specifically designed for bird species recognition. It contains a total of $11{,}788$ images covering $200$ bird species. Each class has approximately $60$ images, although the count varies slightly across classes. The dataset is split into $5{,}994$ training images and $5{,}794$ testing images.

\vspace{10pt}
\noindent \textbf{FGVC-Aircraft}~\cite{maji2013fine} is a fine-grained dataset designed for the visual categorization of aircraft models. It comprises $10{,}000$ images spanning $100$ aircraft classes, where each class corresponds to a specific aircraft model variant (e.g., Boeing 747-400, Airbus A320-200). The dataset is split into $6{,}667$ training, $1{,}333$ validation, and $2{,}000$ test images.

\vspace{10pt}
\noindent \textbf{Stanford-Cars}~\cite{KrauseStarkDengFei-Fei_3DRR2013} is a fine-grained dataset consisting of $16{,}185$ images of cars, split into $8{,}144$ training and $8{,}041$ test images. The dataset includes $196$ car classes, each corresponding to a specific make, model, and year. All images are high resolution and curated from real-world scenarios. The dataset is intended for fine-grained classification, where subtle visual distinctions between very similar car types must be learned.

\section{Implementation Details}
\label{sec:supp_implementation_details}
We follow the protocols and configurations of AdAutoMix~\cite{qinadversarial} for fair comparison with prior SOTA mixup methods. All experiments are conducted within OpenMixup~\cite{li2022openmixup}.

\vspace{10pt}
\noindent \textbf{CIFAR-100.} We apply basic data augmentations consisting of random flip and random crop with 4-pixel padding for $32\times32$ images. For ResNet-18 and ResNeXt-50, the training setup uses the SGD optimizer (momentum $=0.9$, weight decay $=0.0001$), a batch size of $100$, and $200$ training epochs. The initial learning rate is set to $0.1$ and decayed via a cosine scheduler.

\vspace{10pt}
\noindent \textbf{ImageNet-1K.} We adopt a PyTorch-style training configuration, training the model for $100$ epochs using SGD with a batch size of $256$. The initial learning rate is $0.1$, the weight decay is $0.0001$, and the momentum is $0.9$.

\vspace{10pt}
\noindent \textbf{CUB-200, FGVC-Aircraft, and Stanford-Cars.} For all three fine-grained datasets we initialize the models with the official PyTorch pre-trained weights on ImageNet-1K. Training uses SGD (momentum $=0.9$, weight decay $=0.0005$), a batch size of $16$, and $200$ epochs. The learning rate starts at $0.001$ and is adjusted dynamically via a cosine scheduler. The hyperparameters $\lambda$ and $t$ are set to $0.2$ and $0.5$, respectively.

\section{Additional Ablation Studies}
\label{sec:supp_ablations}
This section provides further empirical evidence on the design choices of \our{}. We begin by comparing our framework with existing multi-mode mixing baselines, then examine the contribution of the multi-mode wrapper itself, study the sensitivity of the fractal library size, contrast targeted fractal injection with the global fractal strategy adopted by PixMix, place our method in a conceptual comparison with prior single-image and fractal-based augmentations, present a direct head-to-head with SalfMix, and finally evaluate the downstream behavior of \our{} when fine-tuning large foundation backbones.

\subsection{Comparison with Multi-Mode Mixing Baselines}
\label{sec:multimode_baselines}
A natural concern is whether the gains of \our{} arise from the multi-mode wrapper rather than from the saliency-fractal core. To address this, we compare against RandomMix~\cite{liu2025randomix}, which combines Mixup, CutMix, ResizeMix, and FMix, and against AdAutoMix~\cite{qinadversarial} augmented with the same multi-mode wrapper that we apply to our method (i.e.\ $M_m{+}M_c{+}M_r$). Results on CIFAR-100 with ResNet-18 and ResNeXt-50 are reported in Tab.~\ref{tab:multimode_baselines}.

\begin{table}[h]
\small
\centering
\caption{Comparison with multi-mode mixing baselines on CIFAR-100 using ResNet-18 (Res-18) and ResNeXt-50 (ResXt-50).}
\label{tab:multimode_baselines}
\scalebox{1.0}{
\begin{tabular}{lcc}
\toprule
Method & Res-18 & ResXt-50 \\
\midrule
RandomMix~\cite{liu2025randomix}              & 81.02 & 82.21 \\
AdAutoMix~$+\,M_m{+}M_c{+}M_r$                 & 81.06 & 82.56 \\
\textbf{\our{}} (no multi-mode)               & \textbf{81.73} & \textbf{82.22} \\
\bottomrule
\end{tabular}
}
\end{table}

\our{} delivers superior accuracy on both backbones at substantially lower computational cost. The key takeaway is that the bare saliency-fractal core already outperforms two strong multi-mode baselines, indicating that the benefit of our framework comes primarily from the targeted saliency-guided fractal injection rather than from the act of combining multiple mixing modes.

\subsection{Effect of the Multi-Mode Wrapper on \our{}}
\label{sec:multimode_effect}
Building on the previous observation, we next quantify the additional gain contributed by the high-level multi-mode wrapper when it is layered on top of \our{}. Tab.~\ref{tab:multimode_effect} reports Top-1 accuracy on CIFAR-100, Tiny-ImageNet, and ImageNet-1K across seven backbones, covering both convolutional and transformer architectures.

\begin{table}[h]
\centering
\caption{Effect of the multi-mode wrapper across backbones and datasets. \our{}~(only) denotes our core saliency-fractal augmentation without the high-level multi-mode wrapper, while \our{}{+}Multi-Mode additionally enables the random selection of $M_m$, $M_c$, and $M_r$ alongside our core mechanism. AdAutoMix results follow~\cite{qinadversarial} as reported.}
\label{tab:multimode_effect}
\scalebox{1.0}{
\begin{tabular}{lccccccc}
\toprule
Method & \multicolumn{2}{c}{CIFAR-100} & \multicolumn{2}{c}{Tiny-ImageNet} & \multicolumn{3}{c}{ImageNet-1K} \\
\cmidrule(lr){2-3}\cmidrule(lr){4-5}\cmidrule(lr){6-8}
& Swin-T & CNeXt-T & Res18 & ResXt50 & Res18 & Res34 & Res50 \\
\midrule
AdAutoMix~\cite{qinadversarial}    & 84.33 & 83.54 & 69.19 & 72.89 & 70.86 & 74.82 & 78.04 \\
\our{} (only)                       & 85.04 & 84.22 & 70.17 & 73.87 & 71.27 & 75.28 & 78.48 \\
\textbf{\our{}\,+\,Multi-Mode}      & \textbf{85.35} & \textbf{84.41} & \textbf{70.38} & \textbf{74.27} & \textbf{71.37} & \textbf{75.34} & \textbf{78.54} \\
\bottomrule
\end{tabular}
}
\end{table}

Two trends are visible. First, even without the multi-mode wrapper, \our{} consistently outperforms AdAutoMix across all seven settings, with the largest absolute gains observed on Tiny-ImageNet (up to $0.98\%$). Second, enabling the multi-mode wrapper yields a further but smaller improvement of roughly $0.05\%$ to $0.40\%$, confirming that the wrapper acts as a complementary regularizer rather than as the principal source of improvement. We note that the AdAutoMix numbers reported in Tab.~1 of the main paper already incorporate auxiliary augmentations adopted by~\cite{qinadversarial}, which explains why the comparison against the multi-mode wrapper is most directly informative.

\subsection{Fractal Library Size Sensitivity}
\label{sec:fractal_library}
\begin{figure}[h]
\centering
\includegraphics[width=0.7\linewidth]{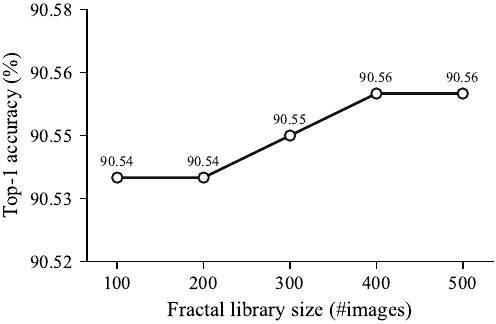}
\caption{Top-1 accuracy on Stanford-Cars when varying the fractal library size from $100$ to $500$ images.}
\label{fig:fractal_lib}
\end{figure}

We next study how the size of the precomputed fractal library affects augmentation quality. A small library risks under-diversifying the perturbation distribution, while an excessively large library increases storage cost without a clear performance benefit. We vary the number of fractal patterns from $100$ to $500$ and report Top-1 accuracy on Stanford-Cars in Fig.~\ref{fig:fractal_lib}.

The curve is essentially flat between $100$ and $200$ fractals ($90.54\%$), rises modestly through $300$ ($90.55\%$), and plateaus from $400$ onward at $90.56\%$. The saturation occurs well before $500$ fractals, which suggests that the role of the library is to provide enough self-similar variation to act as structured noise rather than to densely cover a fractal manifold. We therefore adopt $400$ fractals as the default in all main experiments, which keeps the precomputation cost negligible.

\subsection{Comparison with PixMix on Safety Benchmarks}
\label{sec:pixmix_comparison}
PixMix~\cite{hendrycks2022pixmix} is the closest prior work in the fractal-augmentation family. The key distinction is that PixMix blends fractals across the entire image, whereas \our{} confines fractal injection to the salient regions identified by our self-saliency module. To isolate the effect of targeted injection, we follow the PixMix evaluation protocol on five safety benchmarks. Results are reported in Tab.~\ref{tab:pixmix_safety}.

\begin{table}[h]
\small
\centering
\caption{Safety benchmark comparison with PixMix. Lower is better for corruption, consistency, adversarial, and calibration error, while higher is better for anomaly detection.}
\label{tab:pixmix_safety}
\scalebox{1.0}{
\begin{tabular}{lcccccc}
\toprule
Metric & Mixup & CutMix & AugMix & Outlier & PixMix & \textbf{Ours} \\
\midrule
Corruptions $\downarrow$    & 48.0 & 51.5 & 35.4 & 51.5 & 30.5 & \textbf{27.8} \\
Consistency $\downarrow$    &  9.5 & 12.0 &  6.5 & 11.3 &  5.7 & \textbf{4.8}  \\
Adversaries $\downarrow$    & 97.4 & 97.0 & 95.6 & 97.2 & 92.9 & \textbf{89.2} \\
Calibration $\downarrow$    & 13.0 & 29.3 & 18.8 & 15.2 &  8.1 & \textbf{7.12} \\
Anomaly Det.~$\uparrow$     & 71.7 & 74.4 & 84.9 & 90.3 & 89.3 & \textbf{90.4} \\
\bottomrule
\end{tabular}
}
\end{table}
\our{} achieves the best score on every safety axis, with the largest relative improvement on adversarial robustness ($89.2$ against $92.9$ for PixMix). On the classification task under the PixMix protocol, \our{} attains a Top-1 error of $19.8\%\downarrow$, which is $2.8\%$ lower than the $22.6\%\downarrow$ reported by PixMix. Combined with the fine-grained Stanford-Cars comparison ($92.78\%$ for \our{} against $86.73\%$ for PixMix), these results confirm that constraining fractal perturbations to salient content is consistently more effective than global blending.

\subsection{Conceptual Comparison with Related Methods}
\label{sec:conceptual_comparison}
To position \our{} within the broader landscape of saliency-driven and fractal-based augmentations, Tab.~\ref{tab:conceptual} summarizes the design ingredients of representative prior methods along five axes, namely the use of saliency, single-image (self) operation, fractal injection, targeted application of fractals, and multi-scale patch construction.

\begin{table}[h]
\small
\centering
\caption{Conceptual comparison with representative saliency-based and fractal-based augmentation methods.}
\label{tab:conceptual}
\scalebox{1.0}{
\begin{tabular}{lccccc}
\toprule
Method & Saliency & Self & Fractal & Target & Multi-Scale \\
\midrule
SaliencyMix~\cite{uddin2020saliencymix}     & \ding{51} & \ding{55} & \ding{55} & \ding{55} & \ding{55} \\
PuzzleMix~\cite{kim2020puzzle}              & \ding{51} & \ding{55} & \ding{55} & \ding{55} & \ding{55} \\
Co-Mixup~\cite{kim2021co}                   & \ding{51} & \ding{55} & \ding{55} & \ding{55} & \ding{55} \\
GuidedMixup~\cite{kang2023guidedmixup}      & \ding{51} & \ding{55} & \ding{55} & \ding{55} & \ding{55} \\
SalfMix~\cite{choi2021salfmix}              & \ding{51} & \ding{51} & \ding{55} & \ding{55} & \ding{55} \\
PixMix~\cite{hendrycks2022pixmix}           & \ding{55} & \ding{55} & \ding{51} & \ding{55} & \ding{55} \\
IPMix~\cite{huang2023ipmix}                 & \ding{55} & \ding{55} & \ding{51} & \ding{55} & \ding{55} \\
DiffuseMix~\cite{islam2024diffusemix}       & \ding{55} & \ding{55} & \ding{51} & \ding{55} & \ding{55} \\
\midrule
\textbf{\our{}}                              & \ding{51} & \ding{51} & \ding{51} & \ding{51} & \ding{51} \\
\bottomrule
\end{tabular}
}
\end{table}
Two clusters of prior work are visible. The first comprises saliency-driven approaches such as SaliencyMix, PuzzleMix, Co-Mixup, and GuidedMixup, which operate across two distinct images and therefore introduce semantic interference whenever the source and target images carry incompatible content. SalfMix is conceptually closer to our work since it operates within a single image, but it does not employ fractals and reaches only $78.35\%$ on CIFAR-100 with ResNet-18. The second cluster includes PixMix, IPMix, and DiffuseMix, which use fractal patterns globally over the entire image and therefore disrupt the underlying object semantics. \our{} unifies the strengths of both clusters by combining single-image saliency-guided construction with targeted fractal injection at multiple scales, and is the only method in the table that satisfies all five design criteria simultaneously.

\subsection{Direct Comparison with SalfMix}
\label{sec:salfmix}
Because SalfMix~\cite{choi2021salfmix} shares the single-image self-saliency idea with our method, we report a head-to-head comparison on CIFAR-100 with ResNet-18 under matched training settings. SalfMix reaches $78.35\%$ Top-1 accuracy, whereas \our{} attains $\mathbf{82.74\%}$, a gap of $4.39\%$. The improvement is attributable to two mechanisms missing from SalfMix, namely the multi-scale patch construction that injects scale invariance and the targeted fractal blending that increases structural complexity inside the salient region while preserving the background context.

\subsection{Downstream Fine-tuning with Foundation Backbones}
\label{sec:foundation}
A practical question is whether the benefits of \our{} transfer to large pre-trained foundation backbones. To answer this, we fine-tune CLIP ResNet-50 and DINOv2 ViT-S on CUB-200 and Stanford-Cars with several mixup augmentations applied during the fine-tuning stage. Tab.~\ref{tab:foundation} reports the Top-1 accuracy.

\begin{table}[h]
\small
\centering
\caption{Downstream fine-tuning of CLIP ResNet-50 and DINOv2 ViT-S on CUB-200 and Stanford-Cars under different augmentations.}
\label{tab:foundation}
\scalebox{1.0}{
\begin{tabular}{lcccc}
\toprule
\multirow{2}{*}{Method} & \multicolumn{2}{c}{CLIP ResNet-50} & \multicolumn{2}{c}{DINOv2 ViT-S} \\
\cmidrule(lr){2-3} \cmidrule(lr){4-5}
& CUB-200 & Stan.-Cars & CUB-200 & Stan.-Cars \\
\midrule
No Augmentation          & 79.97 & 87.10 & 88.64 & 88.30 \\
Mixup~\cite{guo2019mixup} & 80.22 & 88.27 & 89.01 & 88.32 \\
CutMix~\cite{yun2019cutmix}& 79.29 & 88.24 & 88.59 & 88.53 \\
\midrule
\textbf{\our{}}          & \textbf{82.47} & \textbf{89.63} & \textbf{88.97} & \textbf{88.68} \\
\bottomrule
\end{tabular}
}
\end{table}
\our{} delivers the best Top-1 accuracy in all four settings. The relative gain is largest for CLIP ResNet-50 on CUB-200, where our method improves by $2.50\%$ over the no-augmentation baseline and by $2.25\%$ over Mixup. The improvement is more modest for DINOv2 ViT-S, which is expected because the DINOv2 representation is already strongly invariant to many of the perturbations introduced by mixup augmentations. The fact that \our{} still produces a consistent gain even on this strong backbone supports the view that targeted saliency-guided fractal injection introduces a form of structured noise that complements, rather than duplicates, the invariances learned during self-supervised pre-training.

\section{Limitations of \our{}}
\label{lab:limitations}
We discuss the limitations of \our{} to provide a balanced view of its applicability. Our proposed \our{} tackles the important question of \textbf{`how many ways to mix?'}, while existing approaches typically consider only one way to mix. Three aspects of the current realization warrant discussion.

\vspace{6pt}
\noindent \textbf{Random selection of mixing modes.} The current realization adopts random selection across high-level mixing modes, which may not exploit task-specific or instance-dependent structure. Replacing random selection with a learned gating mechanism would likely improve performance but would introduce additional training overhead, which conflicts with our explicit goal of a low-cost augmentation framework.

\vspace{6pt}
\noindent \textbf{Manual hyperparameter setting.} All mixup parameters in \our{}, including the saliency threshold $t$ and the fractal blending coefficient $\lambda$, are manually set rather than learned. This choice favors reproducibility and simplifies hyperparameter transfer across datasets and architectures, at the cost of adaptability to instance-specific augmentation budgets.

\vspace{6pt}
\noindent \textbf{Single-modality scope.} The current method targets single-modality visual data and does not natively extend to video or multimodal inputs. Extending the framework to temporal data would require a saliency module that operates over space and time, as well as a notion of fractal injection that respects temporal coherence.

\subsection{Limitation of the Saliency Detector}
\label{sec:saliency_limit}
Since \our{} relies on a spectral residual saliency detector~\cite{hou2007saliency} to identify regions for patch extraction and fractal injection, its behavior depends on the quality of this detector. Our design goal is to remain label-preserving and to avoid generating unrealistic images while maintaining a low memory footprint. When the saliency detector moderately misidentifies salient regions, the overall performance of our method is largely unaffected because the multi-scale patch construction and the multi-mode wrapper compensate for local errors. In the more severe regime, where the detector fails catastrophically, the effectiveness of the self-saliency step diminishes, but the degradation remains graceful since the structured rotation, blur, and fractal blending operations continue to provide useful regularization signal even in the absence of accurate saliency guidance.

\end{document}